\documentclass[accepted]{safeai2026}

\usepackage[american]{babel}

\usepackage{natbib}
\usepackage{mathtools}
\usepackage{booktabs}
\usepackage{graphicx}

\title{A Translational Note on AI Safety Evaluation}

\author[1]{Madhava~Gaikwad}
\affil[1]{%
    Independent Researcher\\
    Bengaluru, India
}

\begin{document}
\maketitle

\begin{abstract}
Recent studies report that automated red-teaming finds more vulnerabilities, at lower cost, than human red-teaming on standard AI safety benchmarks, and some read this as evidence that human evaluators are becoming dispensable. The comparison measures one thing and the conclusion claims another. A benchmark measures how thoroughly an attacker searches a predefined set of harms, fixed in advance by the developers, and a harm left out of that set is invisible to any attacker working inside it, automated or not. The same blind spot appeared in academic cryptography and in clinical drug trials, where an evaluation that was internally valid stayed silent about the population it was never pointed at. We call the AI-safety version the \emph{threat-model coverage gap}, and find that it persists in a current open-weight model, where harms surface in non-English prompts that English benchmarks miss. Closing it requires evaluators whose deployment context differs from the developers'. The case for those evaluators is methodological, grounded in coverage, and the existing evaluation frame is unlikely to produce them on its own.
\end{abstract}

\section{Introduction}\label{sec:intro}

\begin{figure*}[t]
  \centering
  \includegraphics[width=0.98\textwidth]{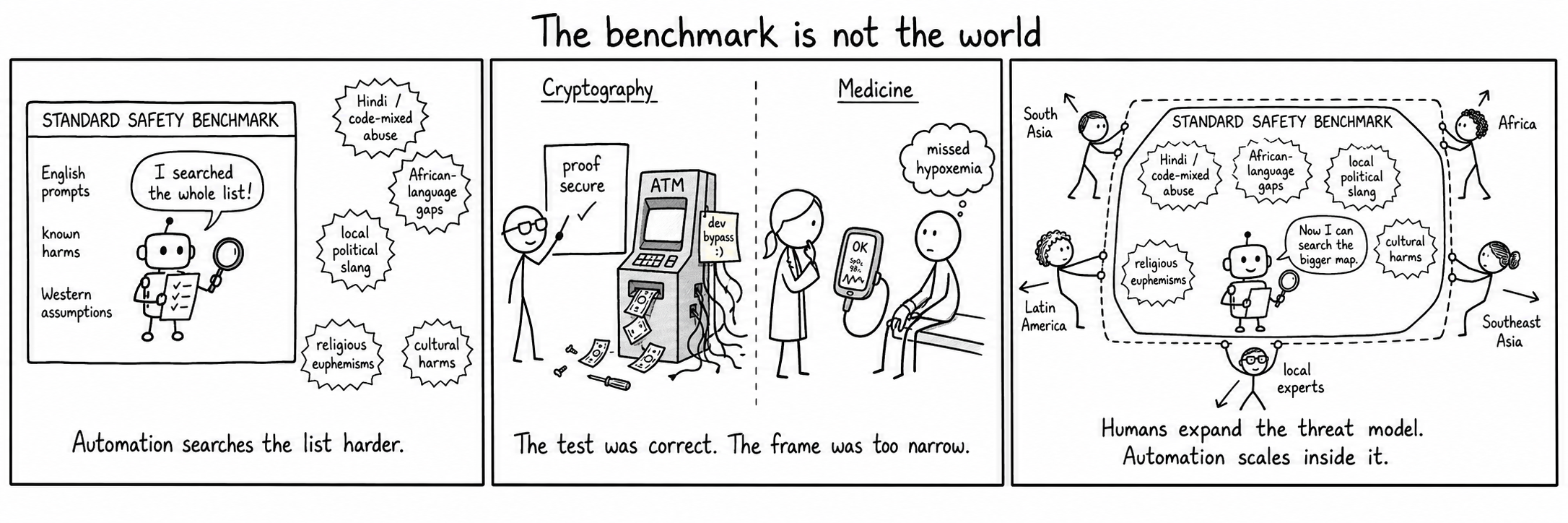}
  \caption{The argument in three panels. Left: automation searches a stated threat model efficiently but never surfaces harms the model omits. Center: the same blind spot in cryptography and medicine, where a valid test was scoped too narrowly. Right: evaluators whose deployment context differs from the developers' enlarge the threat model, after which automation scales the search across the larger space.}
  \label{fig:threat-model-expansion}
\end{figure*}

A growing body of work pits human red-teaming against automated red-teaming on large language models. By red-teaming we mean probing a deployed model with adversarial inputs to find prompts that elicit unsafe outputs. Across benchmarks including Crucible, AIRTBench, and HarmBench, the automated attacker wins. It finds more vulnerabilities, faster and cheaper, than its human counterpart \citep{mulla2025crucible,zhou2025autoredteamer,dawson2025airtbench}. From there it is a short step to the claim that automation is replacing human red-teaming, with the budget implications that follow.

The step looks shorter than it is, because it accepts the benchmark's framing as given. Every benchmark fixes a \emph{threat model}: the harms, attack types, and contexts it instruments. Inside that fixed set, automation does scale better than people. Whether the set matches the harms the system produces once deployed is a different question, and the benchmark cannot answer it. An optimizer pointed at a fixed set searches it harder; it has no way to notice that the set is missing entries.

The omissions are patterned. The standard benchmarks are written in English, by teams clustered in a few jurisdictions, and they carry the assumptions of those jurisdictions. The harms that slip past them concentrate in non-English languages and non-Western deployment contexts \citep{singhania2025mmart,abdullahi2026ubuntuguard,duke2025t2i}. We call the distance between the population a threat model is written for and the population a system is deployed to the \emph{threat-model coverage gap}.

This gap is not peculiar to AI. Cryptography encountered it in the early 1990s, and clinical medicine encountered it across drug trials and pulse oximetry over several decades. In each case the evaluation was internally valid but scoped too narrowly: it said nothing about a deployed population it never examined, and the failure stayed hidden until someone looked from outside the procedure. We call this the internal-validity trap, and it recurs throughout the paper. In every instance the fix came from evaluators outside the original community, not from sharper versions of the existing tests.

\textbf{Contribution.} We bring a methodological pattern from two mature fields into a live AI-safety debate, and do four things with it. First, we map the cryptography episode (\citet{anderson1993cryptosystems}, side channels, fault injection) and the clinical-medicine episode (the exclusion of women from trials, pulse oximetry) onto red-teaming as one recurring structure. Second, we isolate the \emph{threat-model coverage gap} as a concept separate from diversity or multilingual safety, and propose a two-axis evaluation that pulls apart search efficiency and threat-model expansion. Third, we confront the strongest counterargument---that post-deployment patching will absorb the gap---and locate where that channel breaks. Fourth, we supply an empirical anchor cheap enough to run on a consumer laptop, so a practitioner can reproduce it or extend it to other languages, models, or harm taxonomies.

\section{Two Precedents}\label{sec:precedents}

\textbf{Cryptography.} For roughly two decades, academic cryptography judged a system by the security of its underlying mathematics. Provable security and competitive cryptanalysis worked inside that frame, and their proofs held on their own terms. Then \citet{anderson1993cryptosystems} surveyed fraud against retail banking ATMs and found the frame pointed at the wrong target. Although banks ran some of the largest cryptographic deployments of the era, almost all the fraud traced to implementation defects, deployment mistakes, and exploited human procedure. Cryptanalysis of the mathematics accounted for almost none of it. The evaluation had the mathematics right, but the systems broke everywhere the evaluation was not looking.

Anderson traced the gap to who the evaluators were. A mathematical training points attention at mathematical attacks. The failures, though, lived in hardware, software engineering, and organizational practice, where the field knew little. The attack classes that later widened the threat model arrived from outside it. Side channels recover a key from how long an operation takes or how much current it draws \citep{kocher1996timing}. Fault injection breaks a system by provoking hardware errors \citep{boneh1997fault}. These did not come from harder cryptanalysis. They came from researchers whose background covered ground the original community had ruled out of scope.

\textbf{Clinical Medicine.} Drug evaluation repeated the structure in a different field. From 1977 to 1993 the U.S.\ Food and Drug Administration kept women of childbearing potential out of most early-phase trials \citep{fda1977guideline}, a rule first defended as protective, so that for sixteen years dosing, efficacy, and adverse-event data were collected from a mostly male population. When the FDA reversed course, its new guidance acknowledged that drugs vetted on a homogeneous group had, in effect, been tuned to middle-aged white men \citep{fda1993guideline,merkatz1993women}.

Pulse oximetry brings the pattern closer to the present. The devices were calibrated mostly on light-skinned subjects. \citet{sjoding2020} reported in the \emph{New England Journal of Medicine} that Black patients had nearly three times the rate of occult hypoxemia---dangerously low blood oxygen the device reads as normal---compared with white patients. The technology had been in clinical use for decades, and the effect had been documented years earlier without prompting a fix \citep{bickler2005}.

Both medical cases are the internal-validity trap again: the calibration was sound for the population it was drawn from, and useless as a warning about the population it omitted.

\section{The Analogy to AI Safety}\label{sec:analogy}

The same structure appears in AI safety evaluation, and the argument runs in three steps.

\textbf{Claim 1: the threat model carries the fingerprint of whoever built it.} A red-teaming benchmark decides which harms to test, which attacks to consider, and which deployment contexts to instrument, and those decisions track the team that made them. Today that means English, U.S.\ and U.K.\ law, and Western institutional assumptions.

\textbf{Claim 2: a test built from a specification rarely surfaces harms the specification left out.} MM-ART, an automated multilingual method, reports up to 195\% more safety failures in non-English multi-turn conversations than in single-turn English on the same models \citep{singhania2025mmart}. UbuntuGuard, built from queries written by 155 African domain experts, finds that English-centric benchmarks overstate real-world multilingual safety and that cross-lingual transfer gives only partial, insufficient coverage \citep{abdullahi2026ubuntuguard}. In a controlled study of 118 participants from India and the United States, \citet{agarwal2025homogenize} find that Western-centric writing assistants push output toward Western styles and help American users more---the coverage pattern recurring in a non-safety task with native speakers. The STAR framework ties part of the gap directly to who does the red-teaming: evaluators catch the failure modes their own background prepares them to expect \citep{weidinger2024star}. The deployed population is wider than the evaluated one, both in the prompts tested and in the people writing them.

\textbf{Claim 3: automation cannot close the gap from inside the specification.} Automated methods mutate prompts, optimize adversarial inputs, and search the space of known harm categories. They do not infer that the taxonomy itself is short an entry, because spotting a missing category usually takes familiarity with a deployment context where that category bites. Pointing an optimizer harder at an English-centric taxonomy does not make it conclude that the taxonomy is English-centric.

One channel could close the gap without changing who evaluates: post-deployment patching, in which incidents seen in the field feed back into the harm taxonomy, the next training run, or safety fine-tuning. The channel does real work: the gap we measure below is far smaller than the 79\% jailbreak rate \citet{yong2023lowresource} reported two years earlier, which fits incidents from that period having been absorbed. Its limit is at the point of capture, where which incidents reach the pipeline depends on who reports them, in what language, and through which institutional door. The dominant post-deployment record is the incident database, which catalogs real-world failures to feed safety improvements \citep{mcgregor2021aiid}, and such databases are populated mostly from English-language news, so they over-represent Anglophone harms and under-capture harms affecting marginalized or geographically remote populations \citep{allaham2025incidentnews}. A harm whose victims cannot reach that door stays underrepresented at capture, no matter how aggressive the patching that follows. Because the update cycle is itself authored by the developer population, a faster cycle accelerates the coverage that already exists but does not, by itself, extend it. Cryptography drew the same line: automated cryptanalysis improved for decades and produced none of the side-channel or fault-injection classes, because those categories were not reachable from inside the existing reporting frame, however hard the search was pushed.

This reconciles two literatures that look opposed. One finds that automation beats humans, which holds inside a fixed threat model. The other finds that evaluators from underrepresented contexts catch failures automation misses, which is a measure of how far the threat model grows. The defender has the harder side of this. A defender works from a fixed list of harms, while an attacker does not. Translation makes the problem worse, because it drops the signals that carry an attack in another language: slang, religious and political references, code-mixing, local euphemism. Both English-speaking evaluators and English-trained systems lose those signals \citep{shen2024language,peppin2025multilingual}.

\textbf{What would refute this.} The thesis is empirical and makes bets it can lose. Three results would count against it. First, if an automated attacker confined to an English-centric taxonomy surfaced context-specific non-Western harms at the rate evaluators native to those contexts do, Claim 3 would fail, since the coverage gap would be reachable from inside the specification after all. Second, if the harms that evade English benchmarks proved to be randomly distributed across languages and jurisdictions, the premise that omissions are patterned would fail. Third, if incident-reporting pipelines showed no language or access skew and absorbed non-English harms as quickly as English ones, the point-of-capture argument against post-deployment patching would fail. The probe in Section~\ref{sec:empirical} is a first test of the second prediction: had leakage appeared evenly across English and non-English prompts, the patterned-omission premise would have been undercut on the spot.

\section{An Empirical Anchor}\label{sec:empirical}

To test whether the gap survives in a current open-weight model, we ran 100 HarmBench \citep{mazeika2024harmbench} behaviors against \texttt{Llama-3.1-8B-Instruct} in English and five more languages: Hindi, Swahili, Bengali, Yoruba, and Tagalog. We sampled the 100 behaviors stratified across HarmBench's standard, contextual, and copyright subsets in proportion to their registry sizes (50, 25, and 25), drawing uniformly at random within each. NLLB-200 1.3B \citep{nllb2022} did the translation, and Llama-Guard-3-1B \citep{llamaguard3} labeled each response safe or unsafe. The whole pipeline runs locally on a consumer machine (Apple M4 Pro, 24~GB unified memory) in about two hours. We chose the tool stack so the probe reproduces on a student laptop. Code and data are at \url{https://github.com/krimler/harmbench_multilingual/}.

We drop the copyright subset from the reported numbers. It turns out to measure capability instead of safety. The model reproduces English song lyrics on request and cannot do the same in Yoruba, because it cannot write fluent Yoruba lyrics in the first place. We keep the standard and contextual subsets, which the model can attempt in all six languages.

\begin{table}[t]
\centering
\caption{Unsafe-Response Rates on Llama-3.1-8B-Instruct (Standard and Contextual Subsets; Copyright Subset Excluded). Brackets Show 95\% Wilson Confidence Intervals.}\label{tab:results}
\small
\begin{tabular}{lcc}
\toprule
Language & Standard ($n{=}50$) & Contextual ($n{=}25$) \\
\midrule
English & 0\% \,[0.0, 7.1]   & 0\% \,[0.0, 13.3] \\
Hindi   & 4\% \,[1.1, 13.5]  & 20\% \,[8.9, 39.1] \\
Swahili & 4\% \,[1.1, 13.5]  & 24\% \,[11.5, 43.4] \\
Bengali & 8\% \,[3.2, 18.8]  & 16\% \,[6.4, 34.7] \\
Yoruba  & 2\% \,[0.4, 10.5]  & 24\% \,[11.5, 43.4] \\
Tagalog & 4\% \,[1.1, 13.5]  & 12\% \,[4.2, 30.0] \\
\bottomrule
\end{tabular}
\end{table}

Table~\ref{tab:results} reports the result. English drew zero unsafe responses across 75 prompts; the five non-English languages drew 5--8 each across the same 75, for overall rates from 6.7\% (Tagalog) to 10.7\% (Swahili and Bengali). Pooled across the five non-English languages, that is 35 unsafe responses in 375 prompts (9.3\%), against 0 in 75 for English. A one-sided Fisher exact test rejects equal rates at $p = 0.0013$. The per-cell Wilson intervals are wide given the small subsets---on the contextual subset the upper bounds reach 30--43\%---so no single percentage should carry weight. The qualitative result holds regardless: zero leakage in English, non-zero leakage in every non-English language tested, concentrated in the contextual subset. The magnitude sits well below the 79\% jailbreak rate \citet{yong2023lowresource} measured on GPT-4 in late 2023, consistent with two more years of multilingual safety training; what we report is the residual gap that training leaves behind.

\textbf{Limitations.} The experiment is an existence proof. It cannot apportion leakage among translation artifacts, classifier noise, pretraining underexposure, and cultural framing; prior work does that decomposition \citep{abdullahi2026ubuntuguard,shen2024language}. We test one open-weight model, though the cited literature finds the pattern across many \citep{shen2024language,peppin2025multilingual}. The small per-cell sizes (25 contextual, 50 standard) make the intervals wide, which is why we lead with the pooled Fisher test as the strongest defensible claim. As a partial check on translation noise, requiring two-translator chrF agreement above 0.3 moves language-level rates by at most 2.4 points; the matching classifier-noise check is missing and would need Llama-Guard-3-8B or native-speaker annotation. The 24\% Yoruba contextual rate is where that unchecked noise matters most, since it rests on six machine-labeled responses that no Yoruba speaker has verified. And the prompts are machine-translated, so the numbers should be read as a lower bound: \citet{abdullahi2026ubuntuguard} report sharp F1 drops in English-centric guardrails once prompts are fully localized to African languages.

\textbf{Scope.} The probe does not aim to improve on expert-built, culturally grounded benchmarks, which already exist and continue to appear \citep{abdullahi2026ubuntuguard,singhania2025mmart}. It aims at a different constraint: the cost of running a check at all. A two-hour, single-laptop probe that any evaluator can retarget to a new language or model lowers that cost to the point where contributors outside well-funded labs can take part in threat-model expansion, which is what makes the two-axis program in Section~\ref{sec:implications} practical to adopt.

\section{Implications}\label{sec:implications}

\textbf{Implication 1: evaluate on two axes.} Because the human-versus-automation contest folds two activities into one, it should be split. One axis is search efficiency within a stated threat model, where automation is the right tool. The other is threat-model expansion, measured by how fast previously uncatalogued harm categories get surfaced. A corollary: a benchmark release should state its coverage outright---the languages, jurisdictions, and deployment contexts it does and does not represent. A benchmark that does not declare its exclusions cannot be judged on whether they matter.

\textbf{Implication 2: build outside participation in, and keep it there.} Sustained involvement by evaluators outside the developer population is the move Anderson urged for hardware engineers in cryptography and the move the FDA eventually made for women in trials. Neither recovery came from goodwill; both came from new venues, funding, and mandates. The medical case is concrete: the FDA's 1993 \emph{Guideline for the Study and Evaluation of Gender Differences in the Clinical Evaluation of Drugs} \citep{fda1993guideline} turned inclusion into a standing regulatory expectation. The stakes rise with agentic systems, where a coverage gap in evaluation propagates into the actions the system selects. The AI-safety equivalents would make threat-model coverage a standing requirement of benchmark releases and fund non-Western evaluation as a recurring budget line.

\bibliography{refs}

@inproceedings{mulla2025crucible,
  author    = {Mulla, Rob and Dawson, Ads and Abruzzon, Vincent and Greunke, Brian and Landers, Nick and Palm, Brad and Pearce, Will},
  title     = {The Automation Advantage in {AI} Red Teaming},
  booktitle = {arXiv preprint arXiv:2504.19855},
  year      = {2025}
}

@inproceedings{zhou2025autoredteamer,
  author    = {Zhou, Andy and Wu, Kevin and Pinto, Francesco and Chen, Zhaorun and Zeng, Yi and Yang, Yu and Yang, Shuang and Koyejo, Sanmi and Zou, James and Li, Bo},
  title     = {{AutoRedTeamer}: Autonomous Red Teaming with Lifelong Attack Integration},
  booktitle = {arXiv preprint arXiv:2503.15754},
  year      = {2025}
}

@inproceedings{mazeika2024harmbench,
  title={HarmBench: A Standardized Evaluation Framework for Automated Red Teaming and Robust Refusal},
  author={Mazeika, Mantas and Phan, Long and Yin, Xuwang and Zou, Andy and Wang, Zifan and Mu, Norman and Sakhaee, Elham and Li, Nathaniel and Basart, Steven and Li, Bo and others},
  booktitle={Proceedings of the 41st International Conference on Machine Learning ({ICML})},
  series={Proceedings of Machine Learning Research},
  volume={235},
  pages={35181--35224},
  year={2024},
  publisher={PMLR}
}

@article{nllb2022,
  author  = {{NLLB Team} and Costa-juss{\`a}, Marta R. and Cross, James and {\c{C}}elebi, Onur and Elbayad, Maha and Heafield, Kenneth and Heffernan, Kevin and Kalbassi, Elahe and Lam, Janice and Licht, Daniel and Maillard, Jean and Sun, Anna and Wang, Skyler and Wenzek, Guillaume and Youngblood, Al and others},
  title   = {No Language Left Behind: Scaling Human-Centered Machine Translation},
  journal = {arXiv preprint arXiv:2207.04672},
  year    = {2022}
}

@misc{llamaguard3,
  author       = {Meta AI},
  title        = {Llama {G}uard 3 Model Card},
  howpublished = {\url{https://github.com/meta-llama/PurpleLlama/blob/main/Llama-Guard3/8B/MODEL_CARD.md}},
  year         = {2024}
}

@inproceedings{dawson2025airtbench,
  author    = {Dawson, Ads and Mulla, Rob and Landers, Nick and Caldwell, Shane},
  title     = {{AIRTBench}: Measuring Autonomous {AI} Red Teaming Capabilities in Language Models},
  booktitle = {arXiv preprint arXiv:2506.14682},
  year      = {2025}
}

@inproceedings{anderson1993cryptosystems,
  author    = {Anderson, Ross J.},
  title     = {Why Cryptosystems Fail},
  booktitle = {Proceedings of the 1st {ACM} Conference on Computer and Communications Security ({CCS})},
  pages     = {215--227},
  year      = {1993},
  publisher = {ACM}
}

@inproceedings{kocher1996timing,
  author    = {Kocher, Paul C.},
  title     = {Timing Attacks on Implementations of {Diffie-Hellman}, {RSA}, {DSS}, and Other Systems},
  booktitle = {Advances in Cryptology ({CRYPTO})},
  series    = {Lecture Notes in Computer Science},
  volume    = {1109},
  pages     = {104--113},
  year      = {1996},
  publisher = {Springer}
}

@inproceedings{boneh1997fault,
  author    = {Boneh, Dan and DeMillo, Richard A. and Lipton, Richard J.},
  title     = {On the Importance of Checking Cryptographic Protocols for Faults},
  booktitle = {Advances in Cryptology ({EUROCRYPT})},
  series    = {Lecture Notes in Computer Science},
  volume    = {1233},
  pages     = {37--51},
  year      = {1997},
  publisher = {Springer}
}

@techreport{fda1977guideline,
  author      = {{U.S. Food and Drug Administration}},
  title       = {General Considerations for the Clinical Evaluation of Drugs},
  institution = {U.S. Government Printing Office},
  number      = {HEW Publication No. (FDA) 77-3040},
  address     = {Washington, DC},
  year        = {1977}
}

@article{fda1993guideline,
  author  = {{U.S. Food and Drug Administration}},
  title   = {Guideline for the Study and Evaluation of Gender Differences in the Clinical Evaluation of Drugs},
  journal = {Federal Register},
  volume  = {58},
  pages   = {39406--39416},
  year    = {1993}
}

@article{merkatz1993women,
  author  = {Merkatz, Ruth B. and Temple, Robert and Sobel, Solomon and Feiden, Karyn and Kessler, David A.},
  title   = {Women in Clinical Trials of New Drugs: A Change in {Food and Drug Administration} Policy},
  journal = {New England Journal of Medicine},
  volume  = {329},
  number  = {4},
  pages   = {292--296},
  year    = {1993}
}

@article{sjoding2020,
  author  = {Sjoding, Michael W. and Dickson, Robert P. and Iwashyna, Theodore J. and Gay, Steven E. and Valley, Thomas S.},
  title   = {Racial Bias in Pulse Oximetry Measurement},
  journal = {New England Journal of Medicine},
  volume  = {383},
  number  = {25},
  pages   = {2477--2478},
  year    = {2020}
}

@article{bickler2005,
  author  = {Bickler, Philip E. and Feiner, John R. and Severinghaus, John W.},
  title   = {Effects of Skin Pigmentation on Pulse Oximeter Accuracy at Low Saturation},
  journal = {Anesthesiology},
  volume  = {102},
  number  = {4},
  pages   = {715--719},
  year    = {2005}
}

@inproceedings{singhania2025mmart,
  author    = {Singhania, Abhishek and Dupuy, Christophe and Mangale, Shivam and Namboori, Amani},
  title     = {Multi-lingual Multi-turn Automated Red Teaming for {LLMs}},
  booktitle = {Proceedings of the 5th Workshop on Trustworthy {NLP} ({TrustNLP} 2025)},
  pages     = {141--154},
  year      = {2025},
  publisher = {Association for Computational Linguistics}
}

@article{abdullahi2026ubuntuguard,
  author    = {Abdullahi, Tassallah and Mgonzo, Macton and Oduwole, Mardiyyah and Okewunmi, Paul and Owodunni, Abraham and Singh, Ritambhara and Eickhoff, Carsten},
  title     = {{UbuntuGuard}: A Culturally-Grounded Policy Benchmark for Equitable {AI} Safety in {African} Languages},
  journal   = {arXiv preprint arXiv:2601.12696},
  year      = {2026}
}

@inproceedings{weidinger2024star,
  author    = {Weidinger, Laura and Mellor, John and Pegueroles, Bernat Guill{\'e}n and Marchal, Nahema and Kumar, Ravin and Lum, Kristian and Akbulut, Canfer and Diaz, Mark and Bergman, Stevie and Rodriguez, Mikel and Rieser, Verena and Isaac, William},
  title     = {{STAR}: {SocioTechnical} Approach to Red Teaming Language Models},
  booktitle = {Proceedings of the 2024 Conference on Empirical Methods in Natural Language Processing ({EMNLP})},
  pages     = {21516--21532},
  year      = {2024}
}

@misc{duke2025t2i,
  author       = {{Duke MIDS / Meta capstone project}},
  title        = {Blind Spots in Text-to-Image Models: Evaluating Cultural Gaps, Failures, and Hidden Risks in {GenAI} Media Safety},
  howpublished = {Duke University Master in Interdisciplinary Data Science},
  year         = {2025}
}

@inproceedings{yong2023lowresource,
  author    = {Yong, Zheng-Xin and Menghini, Cristina and Bach, Stephen H.},
  title     = {Low-Resource Languages Jailbreak {GPT-4}},
  booktitle = {NeurIPS Workshop on Socially Responsible Language Modelling Research (SoLaR)},
  year      = {2023},
  note      = {arXiv:2310.02446}
}

@inproceedings{shen2024language,
  author    = {Shen, Lingfeng and Tan, Weiting and Chen, Sihao and Chen, Yunmo and Zhang, Jingyu and Xu, Haoran and Zheng, Boyuan and Koehn, Philipp and Khashabi, Daniel},
  title     = {The Language Barrier: Dissecting Safety Challenges of {LLMs} in Multilingual Contexts},
  booktitle = {Findings of the Association for Computational Linguistics: ACL 2024},
  pages     = {2668--2680},
  year      = {2024}
}

@article{peppin2025multilingual,
  author  = {Peppin, Aidan and Kreutzer, Julia and Schoenauer Sebag, Alice and Marchisio, Kelly and Ermis, Beyza and Dang, John and Cahyawijaya, Samuel and Singh, Shivalika and Goldfarb-Tarrant, Seraphina and Aryabumi, Viraat and Aakanksha and Ko, Wei-Yin and {\"U}st{\"u}n, Ahmet and Gall{\'e}, Matthias and Fadaee, Marzieh and Hooker, Sara},
  title   = {The Multilingual Divide and Its Impact on Global {AI} Safety},
  journal = {arXiv preprint arXiv:2505.21344},
  year    = {2025}
}

@inproceedings{agarwal2025homogenize,
  author    = {Agarwal, Dhruv and Naaman, Mor and Vashistha, Aditya},
  title     = {{AI} Suggestions Homogenize Writing Toward {Western} Styles and Diminish Cultural Nuances},
  booktitle = {Proceedings of the {CHI} Conference on Human Factors in Computing Systems ({CHI} '25)},
  year      = {2025},
  publisher = {ACM},
  doi       = {10.1145/3706598.3713564}
}

@inproceedings{mcgregor2021aiid,
  author    = {McGregor, Sean},
  title     = {Preventing Repeated Real World {AI} Failures by Cataloging Incidents: The {AI} Incident Database},
  booktitle = {Proceedings of the {AAAI} Conference on Artificial Intelligence},
  volume    = {35},
  pages     = {15458--15463},
  year      = {2021}
}

@article{allaham2025incidentnews,
  author    = {Allaham, Mowafak and Kieslich, Kimon and Diakopoulos, Nicholas},
  title     = {Global Perspectives of {AI} Risks and Harms: Analyzing the Negative Impacts of {AI} Technologies as Prioritized by News Media},
  journal   = {arXiv preprint arXiv:2501.14040},
  year      = {2025}
}

\end{document}